\documentclass{article}
\usepackage{arxiv}

\usepackage[utf8]{inputenc} 
\usepackage[T1]{fontenc}    
\usepackage{hyperref}       
\usepackage{url}            
\usepackage{booktabs}       
\usepackage{amsfonts}       
\usepackage{nicefrac}       
\usepackage{microtype}      
\usepackage{subfiles}
\usepackage{amsmath}
\usepackage{enumitem}
\usepackage{multirow}
\usepackage{xcolor}
\usepackage{soul}
\usepackage{subfiles}
\usepackage{graphicx}
\usepackage{amssymb}

\usepackage{lineno}
\usepackage[T1]{fontenc}
\usepackage{amsmath}
\usepackage{enumitem}
\usepackage{multirow}
\usepackage{xcolor}
\usepackage{soul}
\usepackage{subfiles}
\title{W2WNet: a two-module probabilistic Convolutional Neural Network with embedded data cleansing functionality}

\author{
  Francesco Ponzio, Enrico Macii, Elisa Ficarra, Santa Di Cataldo\\
  Politecnico di Torino, 10129 Torino, Italy\\
  \texttt{\{name.surname\}@polito.it} \\
}

\begin{document}
\maketitle

\begin{abstract}
Convolutional Neural Networks (CNNs) are supposed to be fed with only high-quality annotated datasets. Nonetheless, in many real-world scenarios, such high quality is very hard to obtain, and datasets may be affected by any sort of image degradation and mislabelling issues. This negatively impacts the performance of standard CNNs, both during the training and the inference phase. To address this issue we propose Wise2WipedNet (W2WNet), a new two-module Convolutional Neural Network, where a \emph{Wise} module exploits Bayesian inference to identify and discard spurious images during the training, and a \emph{Wiped} module takes care of the final classification, while broadcasting information on the prediction confidence at inference time. The goodness of our solution is demonstrated on a number of public benchmarks addressing different image classification tasks, as well as on a real-world case study on histological image analysis. Overall, our experiments demonstrate that W2WNet is able to identify image degradation and mislabelling issues both at training and at inference time, with positive impact on the final classification accuracy.
\end{abstract}


\section{Introduction}
\label{sec:intro}
Since the milestone study by Alex Krizhevsky and colleagues in 2012~\cite{alexnet}, Deep Learning (DL), with particular emphasis on Convolutional Neural Networks (CNNs), is the state-of-the-art method for image classification in many different applications. Besides classification performance, the reason for the success of CNNs is twofold: i)~the recent boost of graphical processing units (GPUs) and parallel processing, that allows to train very large models; ii)~the ever-growing availability of massive annotated task-specific datasets. Nonetheless, in many realistic applications many concerns may be raised about the reliability of such datasets both in terms of image and labelling quality, and consequently on the robustness of the CNN models trained and tested on them. 

As regards to image quality, standard CNNs are supposed to be fed with high quality samples. Nevertheless, in practical scenarios different kinds of image degradation can heavily affect the performance of a CNN both in the training and in the inference phase. Problems concerning image acquisition devices, poor image sensor, lighting conditions, focus, stabilization, exposure time or partial occlusion of the cameras may lead to produce low quality samples, which have been demonstrated to be one of the chief reasons for troublesome learning procedures of CNN models in many applications~\cite{roy2018effects, moosavi2016deepfool, dodge2016understanding}. On the other hand, even though the CNN had been proficiently trained and validated on high quality data, noisy inputs can heavily affect the inference phase, and cause classification errors at run-time. A typical example are self-driving cars, where a partial occlusion of the image acquisition device may lead to misinterpret a road sign, with catastrophic consequences. In such settings, the well-known limitations of standard CNNs to broadcast information about how much the given input resembles the ones the model was trained on - and hence, whether the associated prediction should (or should not) be trusted - is also playing a major role.


Besides image quality, also collecting and associating error-free labels to a massive number of representative images to adequately train CNNs may be extremely problematic in a number of real-world applications. If we take as an example the medical domain, where available data is typically small to begin with, image annotation is always a cumbersome and time-consuming task, that is extremely error-prone. In a number of applications, inter-observer variability is even so high as to necessitate consensus strategies to aggregate annotations from several medical experts~\cite{dataset_labelling1}, which is anyway prone to mislabelling. Conversely, in a number of non-medical real-world scenarios the collection of massive labelled image datasets is relatively easy and straightforward:  for example, using semi-automatic tools based on web search engines and keywords~\cite{dataset_labelling2}. Nonetheless, even in this case concerns may be raised on the reliability of the image labels. Take as an example the JFT dataset from Google, including 300$M+$ images labeled by an algorithm that uses complex mixture of raw web signals, connections between web-pages and user feedback~\cite{Distilling_Hilton, Chollet}: JFT annotations have been found to be 20\% wrong, even after some cleansing procedures~\cite{sun2017revisiting}.

In the rest of this paper, we will refer to image degradation and to mislabelling errors respectively by the name of \emph{measurement} and \emph{labelling noise}.

Even though recent studies have proposed many techniques to compensate the learning degradation due to \emph{measurement noise}~\cite{dodge2016understanding, roy2018effects, moosavi2016deepfool} or \emph{labelling noise} ~\cite{dataset_labelling1, xiao2015learning, sun2017revisiting} specifically, very few researchers have developed solutions to mitigate the impact of generic noise, where the two effects may even coexist. Furthermore, there is still very little scientific understanding of how a CNN may behave in presence of noisy inputs at inference phase, i.e. when the final model is applied to a given application, and how to make a CNN model robust to unpredictable noise effects that may make the inputs considerably different to what the model was specifically trained on. 

In our study, we want to focus the attention on data-perturbation irrespective of whether it is a \emph{measurement} or a \emph{labelling noise}, and we will refer to \emph{spurious} (vs. \emph{meaningful}) samples to indicate images affected by any of the two types of noise.  

We therefore propose \textit{Wise2WipedNet} (\textit{W2WNet}), a CNN-based architecture able to i) model the distribution of spurious samples in a generic dataset, which may be corrupted by both \emph{labelling} and \emph{measurement} noise; ii) clearly identify the spurious samples within the training, by virtue of an adaptive pruning criterion that is fully embedded into the learning algorithm, and focus the training on the only meaningful ones; 
and iii) at inference time, classify never seen images into the learned categories plus one, clearly identifying noisy inputs by means of a statistically sound measure of prediction confidence (see figure \ref{fig:system_architecture_test}).

Hence, our solution exploits the concept of prediction confidence in two ways: (i)~during the training phase, to establish a separability criterion between the good quality (a.k.a. meaningful) and the spurious samples, that is embedded into the learning algorithm to make the network able to focus on the only meaningful ones; and (ii)~during the inference phase, to improve the robustness of the model to ambiguous inputs.

To assess the goodness of our approach in different types of settings, we evaluate \emph{W2WNet} on several state-of-the-art public benchmarks, addressing different image classification tasks and types of noise.  In addition to that, we also provide a real-world case study from the medical imaging domain.  

The rest of the manuscript is structured as follows. In Section~\ref{sec:back} we provide the background and state of the art of our work, and highlight our main contributions. In Section~\ref{sec:method} we describe our proposed methodology and implementation details. In Section~\ref{sec:results} we provide and discuss experimental results, respectively on the public benchmarks and on the real-world case study. Finally, Section~\ref{sec:discussion} provides our final considerations and concludes the paper.
 
 



\section{Background}
\label{sec:back}
As discussed in Section~\ref{sec:intro}, in many real-world cases it is not so obvious to have high quality images to train a CNN with. Most likely, the network will face many issues arising from artifacts during image acquisition, transmission, or storage. This typically affects the training procedure, resulting in a degradation of the model performance \cite{dodge2016understanding, roy2018effects}. Thus, a considerable amount of literature has been published on learning CNNs with low quality images and noisy datasets. In surveillance applications, for instance, face recognition from low quality images is a key aspect, and many studies address learning low-quality faces \cite{face1, face2}. In \cite{ullman2016atoms} the authors show that CNNs behave very differently than human vision system (HVS) in handling minimal recognizable configurations (MIRCs), that is the smallest crop of an input image for which a human observer is able to provide a categorization. More specifically, standard CNNs are generally worse than humans at handling MIRCs, which are typically very small, and hence blurry and low resolved. In \cite{dodge2016understanding}, the authors present the first large scale evaluation of deep networks on natural images affected by different types and different levels of image quality degradation. They show that the existing models are especially vulnerable to blur and noise. Finally, in \cite{roy2018effects}, authors show the effects of degradation on different CNN models, proposing a network setup able to reduce the impact of specific type of perturbations. 

As already discussed, besides \emph{measurement} noise, also manual mislabelling or faulty automatic annotations may lead to unwieldy learning and lower classification performance~\cite{dataset_labelling1, dataset_labelling2}. Previous studies specifically addressing \emph{labelling} noise can be categorized into three main groups:
\begin{enumerate}[label=(\roman*)]
    \item Methods that focus on model selection or design. These methods aim at selecting the model, loss function and training procedures that are most robust to mislabelling~\cite{dataset_labelling1}. Literature shows that most supervised loss functions are not fully robust to faulty labels~\cite{bartlett2006convexity}, unless they are handled by overfitting avoidance~\cite{Survey_label_noise, dataset_labelling2}.
    \item Data cleansing methods. The rationale is in this case to remove samples with incorrect labels. In this sense, voting among an ensemble of classifiers has been proven effective~\cite{dataset_labelling1}. Other strategies include identifying mislabeled instances based on their impact on the training process. For example, \cite{kohler2019uncertainty} prune and re-label training instances by setting a threshold on the classification uncertainty, based on Monte-Carlo (MC) dropout. The challenge of this group of methods is to distinguish the informative samples from the harmful mislabeled ones~\cite{dataset_labelling2}. In this sense, cleansing methods built on top of an uncertainty measure are known to be highly dependent on the given application (i.e. type and level of noise) and even on the architecture of the classifier~\cite{kohler2019uncertainty, dataset_labelling1}. For instance, \cite{kohler2019uncertainty} set a fixed threshold on the uncertainty distribution retrieved from training samples, without modeling the distributions of the uncertainties of the noisy and clean images. Hence, the optimal threshold needs to be tailored to the given application, which may limit the usability in real-world scenarios.
    \item Methods that propose classifier training and labelling noise modeling into a unified framework. This category somehow integrates the two aforementioned families. For instance, probabilistic models have been exploited to model the labelling noise and thereby improve classifiers \cite{kohler2019uncertainty}. Other methodologies aim at identifying and penalizing samples with incorrect labels during the training procedure~\cite{dataset_labelling1}.
\end{enumerate}
While there is a large body of literature coping with either \emph{measurement} or \emph{labelling} noise individually, very little efforts have been directed so far to handling both the problems at one time. Nonetheless, this is a non-trivial issue in most real-world applications, where a-priori knowledge about the type of noise affecting the data may not be available. Moreover, while \emph{labelling} noise affects the only training phase, as the supervised learning requires an appropriate labelling of the training samples, \emph{measurement} noise may affect CNNs even at the inference phase. As already mentioned in Section~\ref{sec:intro}, this may leads standard CNNs to catastrophic failure in several real-world applications. Starting from these considerations, we propose a methodology (a.k.a. \emph{W2WNet}) able on one hand to deal with both \emph{measurement} and \emph{labelling} noise, and, on the other hand, to provide a statistically sound measure of prediction confidence at inference phase. Our methodology follows in the footsteps of the earlier work by~\cite{kohler2019uncertainty}, where the authors exploit uncertainty measures retrieved by MC dropout to identify and remove mislabelled samples. Nevertheless, we are substantially different from~\cite{kohler2019uncertainty} in the following: (i)~we tackle both \emph{measurement} and \emph{labelling} noise in parallel; (ii) we propose an end-to-end framework, embedded into a single CNN model; (iii) we provide a pruning strategy for the spurious samples which is totally automatized and adaptive to the given application; (iv) we exploit prediction uncertainty in two different ways. First, to model, recognize and remove the spurious samples from the training strategy. Second, to broadcast information on the prediction confidence, which is exploited to make CNNs robust to noisy inputs at inference time.

\section{Methods}
\label{sec:method}
\begin{figure*}[!t]
    \centering
    \includegraphics[width=0.9\textwidth]{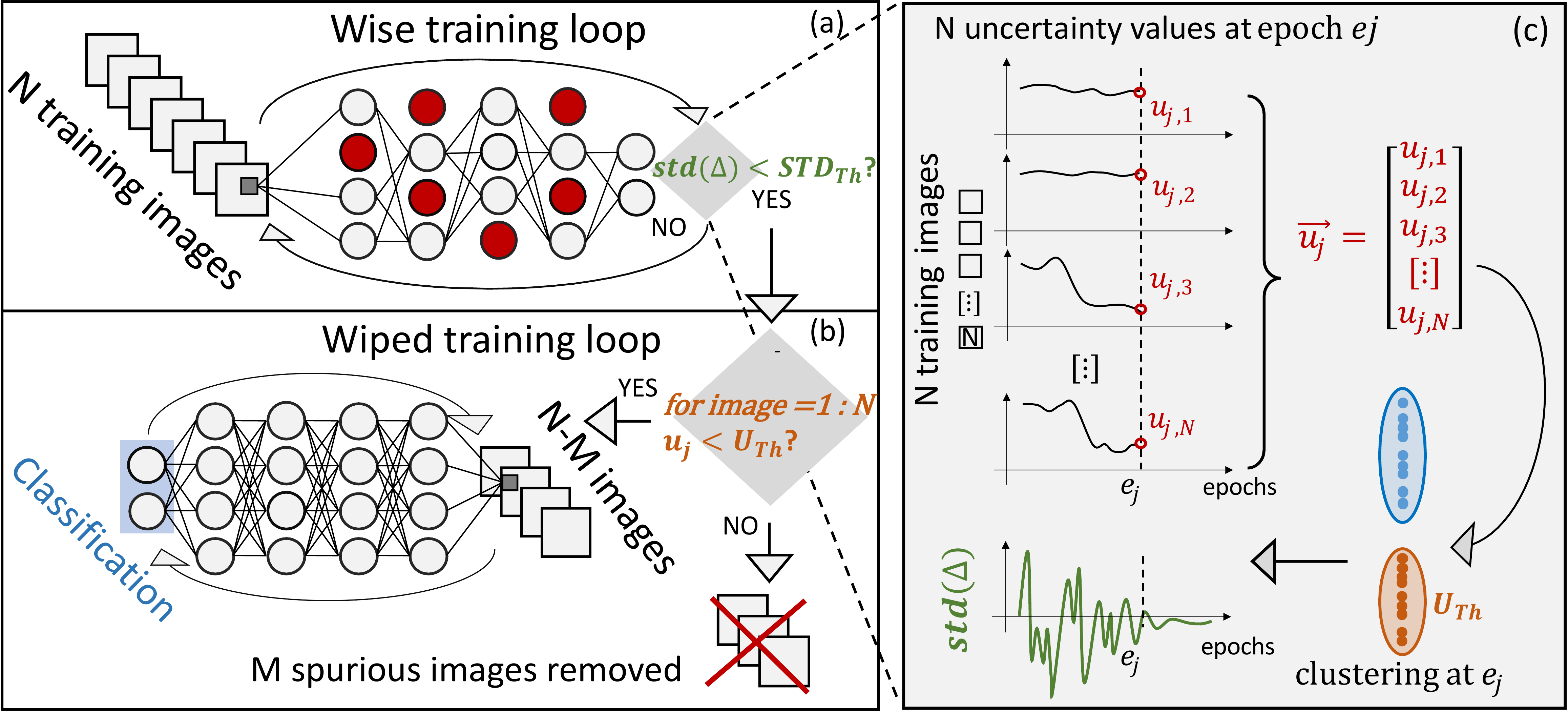}
    \caption{Overview of the training phase of the proposed architecture.}
    \label{fig:system_architecture_train}
\end{figure*}

As represented in \figurename~\ref{fig:system_architecture_train}, our architecture includes two main modules:
\begin{enumerate}[label=(\roman*)]
    \item the \textit{Wise}, that is in charge of a two-fold aim: on one hand, to provide a reliable measure of predictive uncertainty associated to samples (\figurename~\ref{fig:system_architecture_train}(a)); on the other hand, to model the distribution of the spurious samples for the purpose of removing them from the training dataset (\figurename~\ref{fig:system_architecture_train}(c)).
    \item the \textit{Wiped}, that is the expert system trained on the cleaned dataset and designated to the actual classification phase (\figurename~\ref{fig:system_architecture_train}(b)).
\end{enumerate}

\subsection{The Wise: uncertainty estimation}
\label{sec:uncertainty_acquisition}
As already mentioned, the \emph{Wise} must be a noise-aware model, able to associate to each prediction a corresponding uncertainty measure. 
Last trends in deep learning show a growing body of literature around the theme of uncertainty estimation for predictive classification models \cite{lakshminarayanan2017simple, gal2016dropout, kohler2019uncertainty}. With special regards to CNNs, the canonical \textit{softmax} score is erroneously regarded as a measure of prediction confidence, that is: the lower the output of the softmax, the higher the uncertainty on the corresponding prediction. Nonetheless, it has been shown that this is not true, as the softmax merely acts as a normalization \cite{gal2016dropout, hendrycks2016baseline}. As a consequence, a traditional CNN might provide confident (wrong) predictions even on samples that are completely unrelated to what it was specifically trained for.

The most consolidated way to incorporate uncertainty estimation into a CNN leverages on Bayesian formalism~\cite{gal2016dropout, kwon2020uncertainty}. In a Bayesian perspective, individual parameters values (i.e. the weights of the network) are replaced with prior distributions. Hence, the learning strategy is conceived as a probabilistic optimization problem, where the posterior distribution over the parameters is computed, given the training data. As a consequence, the output of the model will also be a posterior predictive distribution of values, from which a statistic can be derived to serve as uncertainty measure. 

Formally, the weights $\omega$ of a CNN are handled as random variables, and assuming the CNN to be exhaustively described by its weights $\omega$, we can write the predictive distribution for a new input $x^*$ as~\cite{gal2016dropout, kwon2020uncertainty}:

\begin{equation}\label{eq:1}
    p(y^*|x^*,X,Y)=\int_{\Omega}p(y^*|x^*,\omega)p(\omega|X,Y)d\omega,
\end{equation} 

Since the term $p(\omega|X,Y)$, integrated upon the whole parameters space $\Omega$, makes the predictive posterior of a CNN analytically and numerically intractable~\cite{lakshminarayanan2017simple, kwon2020uncertainty}, a variety of approximations have been proposed, including Laplace approximation~\cite{laplaceapprox}, Markov chain Monte Carlo (MCMC) methods~\cite{MCMC} and variational Bayesian methods~\cite{variational1, variational2}. Nevertheless, the reliability of the uncertainty measure derived from these approximation strategies strictly depends on two different factors: (i)~the approximation quality constrained by computational requirements; (ii)~the choice of the Bayesian prior, which can ultimately lead to biased predictive uncertainties~\cite{lakshminarayanan2017simple}. In practical terms, Bayesian CNNs (BCNNs) are cumbersome to implement and hard to train, as they require a specific training pipeline handling a very high number of possible hyper-parameters, as well as the high computational cost of the approximation technique~\cite{lakshminarayanan2017simple}.
An interesting insight by Gal and Ghahramani~\cite{gal2016dropout} suggested using Monte Carlo dropout (MC dropout) to estimate predictive uncertainty, which is based on using Dropout~\cite{srivastava2014dropout} at inference time. Since many different neurons are randomly dropped across different model calls, MC dropout method implements a Bayesian sampling from a variational distribution of models. In other words, MC dropout can be seen as an ensemble methodology, where the predictions are averaged over an ensemble of CNNs sharing the same parameters. In such setting, estimating the model uncertainty for a given sample is as simple as keeping the dropout mechanism switched on at inference time, and performing multiple predictions for the same input \cite{lakshminarayanan2017simple}. By using MC dropout, we can then rewrite equation (\ref{eq:1}) with the following approximation:

\begin{equation}\label{eq:2}
    p(y^*|x^*,X,Y) \approx \int_{\Omega}p(y^*|x^*,\omega)q(\omega)d\omega
    \approx \frac{1}{T}\sum_{t=1}^{T}p(y^*|x^*, \hat{\omega}),
\end{equation} 

Thanks to variational inference \cite{gal2016dropout, rkaczkowski2019ara}, we can approximate the posterior distribution $p(\omega|X,Y)$ in (\ref{eq:1}) with a variational one $q(\omega)$. Hence, by means of MC dropout, we assume $q(\omega) \sim \hat{\omega}$, where $\hat{\omega}$ is an estimation resulting from a variational dropout call. 

Starting from the above-mentioned considerations, our \textit{Wise} module (see figure \ref{fig:system_architecture_train}(a)) was implemented as a BCNN, leveraging MC dropout. As anticipated in the previous section, the initial task of the \textit{Wise} is to provide an uncertainty measure for the inputs samples, on the top of which the model can distinguish the spurious samples from the meaningful ones. Downstream of the uncertainty estimation, the \textit{Wise} is able to: (i) identify and eventually remove the spurious samples, thus providing a \emph{cleaned} dataset to train the \textit{Wiped}; (ii) associate a confidence measure to the outcome of the \emph{Wiped}'s classification, that can be exploited to express the reliability of the model's prediction on a given input.

To build our BCNN-based \textit{Wise}, we put into effect equation (\ref{eq:2}) through a DenseNet121 model~\cite{densenet}, inserting a dropout layer with 0.3 rate after each convolutional, pooling and fully connected layer. The DenseNet-based architectures connect all layers directly with each other: each layer obtains additional inputs from all preceding layers and forwards on its own feature-maps to all downstream ones~ \cite{densenet}. By exploiting such \emph{feature reuse} paradigm, DenseNets typically offer exceptional classification capabilities with reduced number of parameters. As it was recently observed that models with less parameters are generally more resilient to image degradations \cite{roy2018effects}, we chose DenseNet121 as best trade-off between classification performances and model compactness. Nonetheless, our \emph{Wise} module can be easily converted into any other state-of-the-art CNN architecture, by simply exploiting MC dropout instead of softmax. 

Before being fed to our model, which is random initialized, samples are pre-processed by zero-centered normalization. The \textit{Wise} is trained with Stochastic Gradient Descent (SGD), setting weight decay to 0.001. 
The number of training epochs of the \textit{Wise} model is a key parameter, which is self-optimized as explained in the next section.

Ultimately, we need to define a statistically sound measure of uncertainty. To do so, we adopted the methodology proposed by Kwon and colleagues \cite{kwon2020uncertainty}: starting from (\ref{eq:2}), the predictive uncertainty of a BCNN may be computed as the sum of the predictive variances of each class [19]. Such predictive variance can be further decomposed into the aleatoric component, able to represent the intrinsic noise in the samples, and the epistemic component, which stems from the parameters and the architecture of the model:

\begin{equation}
    \label{eq:3}
    \underbrace{\frac{1}{T}\sum^T_{t=1}diag(\hat{p}_t)-\hat{p}^{\otimes2}_t}_\text{aleatoric} + \underbrace{\frac{1}{T}\sum^T_{t=1}(\hat{p}_t-\bar{p})^{\otimes2}}_\text{epistemic}
\end{equation}

Here $\bar{p}=\sum^T_{t=1}\hat{p}_t/T$;  $\hat{p}=Softmax{f(\omega_t, x^*)}$ and $T$ is the number of forward passes for input $x^*$. $T$ has been empirically set to $100$ as the best trade-off between computational time and reliability, as stated in \cite{ponzio}.

\subsection{The Wise: modelling of spurious samples distribution}
\label{sec:spurious_modelling}
The aforementioned uncertainty measure provides a way to distinguish between spurious and meaningful samples. The \textit{Wise} has a two-fold functionality. On one hand, during the training phase (see \figurename~\ref{fig:system_architecture_train}), it should identify an epoch $e_{j}$ so that the uncertainty of the spurious samples is significantly higher than the uncertainty of the meaningful samples. Hence, \textit{Wise}'s training should proceed until (i)~the separation between high uncertainty (i.e. spurious) and low uncertainty (i.e. meaningful) samples is large enough, and (ii)~this separation is sufficiently stable over the training epochs. On the other hand, the \textit{Wise} must identify an uncertainty threshold $U_{Th}$ (see \figurename~\ref{fig:system_architecture_train}(c)) that will be exploited at inference time, to broadcast information on the level of confidence of the final prediction (see \figurename~\ref{fig:system_architecture_test}). 

To pursue the stated goals, for a generic $j-th$ training epoch the learning proceeds as follows: 
\begin{enumerate}[label=(\roman*)]
    \item The \textit{Wise} computes for each training sample a corresponding classification uncertainty value, by means of equation (\ref{eq:3}); thus, given $N$ training samples, we obtain a vector of $N$ uncertainty values, referred to as $\Vec{u_j}$ in \figurename~\ref{fig:system_architecture_train}(c); 
    \item The vector $\Vec{u_j}$ is given as input to a K-means clustering, with $K=2$, where the low-uncertainty and the high-uncertainty clusters should represent the meaningful and spurious clusters, respectively. After doing so, the difference between the two clusters' sizes is computed and normalized upon the total number of training samples. Hence, after $j$ training epochs, we obtain a signal $\delta$ made of $j$ such values, whose evolution over time can be exploited to estimate the stability of the clustering at the given epoch. That is, the more stable $\delta$ is over the epochs, the lower the number of samples that are re-assigned to a different cluster, and hence, the more stable the clustering;
    \item At this stage, we need a quantitative stability criterion to stop the \emph{Wise}'s training. First, $\delta$ is low-pass filtered via a median filter with a window size of 11. Second, the standard deviation is computed over a sliding window of size 40 and a stride of 1, obtaining the signal referred to as $std(\Delta)$ at the bottom of \figurename~\ref{fig:system_architecture_train}(c)). To decide on the stability of the clustering at epoch $e_j$, and hence on whether to stop the training, $std(\Delta)$ is imposed a threshold $STD_{Th}$, which is set to 0.01 (see \figurename~\ref{fig:system_architecture_train} (a)). In other words, we stop the training of the \textit{Wise} if more than 99\% of the training samples are stably assigned to the same cluster for 40 consecutive epochs. At inference time, the centroid of the spurious cluster will be exploited as an uncertainty threshold, referred to as $U_{Th}$, in order to identify the samples upon which the model's prediction is not sufficiently confident.  
\end{enumerate} 

\subsection{The Wise \& the Wiped: classification}
\label{sec:wiped}
\begin{figure*}[!t]
    \centering
    \includegraphics[width=0.5\textwidth]{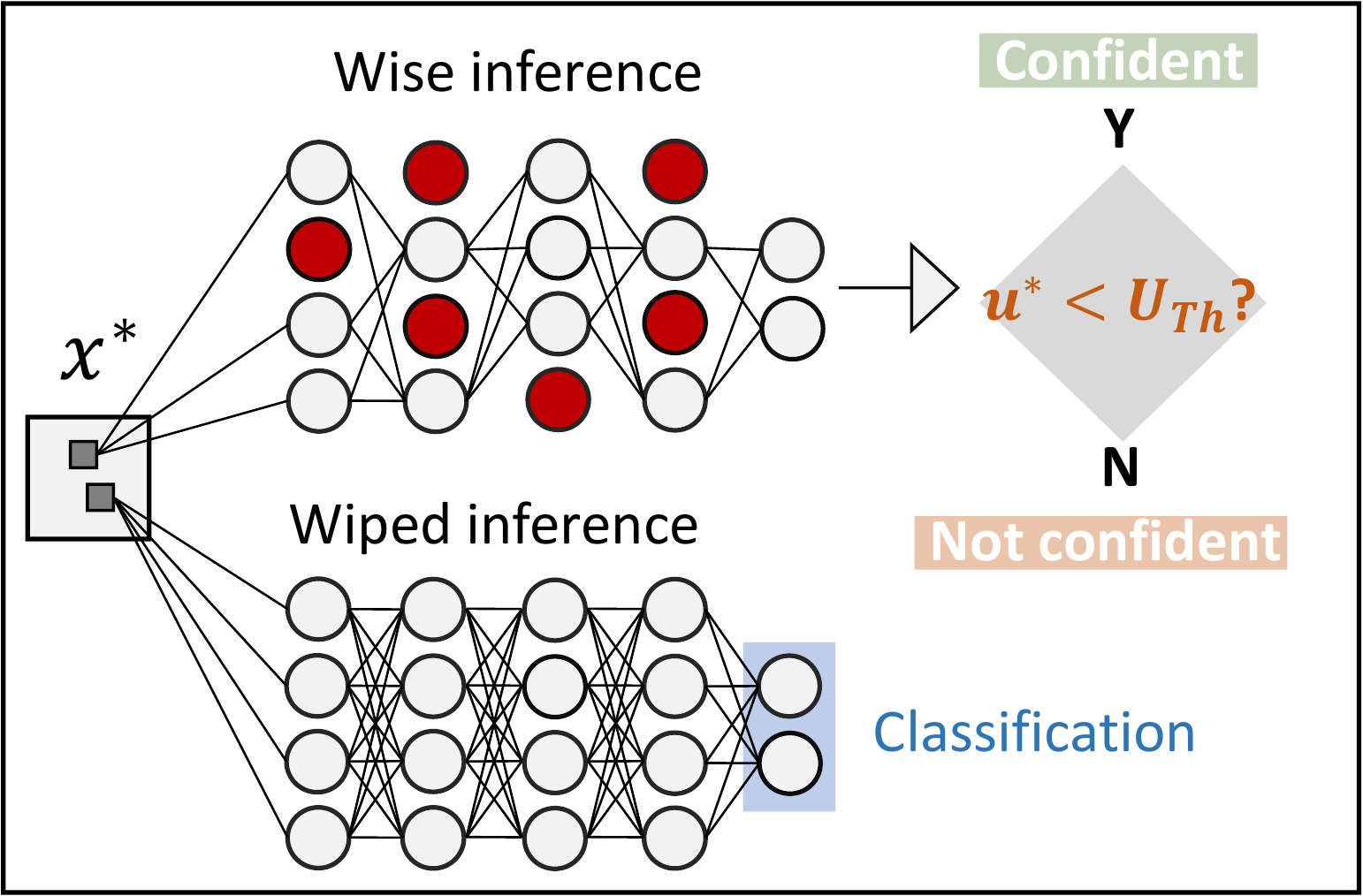}
    \caption{Overview of the inference phase of the proposed architecture.}
    \label{fig:system_architecture_test}
\end{figure*}

While providing a framework to estimate prediction uncertainty, standard BCNNs are often less accurate than their deterministic counterparts at inference time~\cite{shridhar2019comprehensive, ponzio}. To address this issue, as it can be gathered from \figurename~\ref{fig:system_architecture_test}, in our model both the \textit{Wise} and the \textit{Wiped} take part in the inference phase. Given a classification task involving $C$ classes and a generic test sample $x^{*}$, the \textit{Wise} initially computes the corresponding uncertainty $u^*$ through equation \ref{eq:3}. Then, $u^*$ is compared with the threshold $U_{Th}$, identifying $x^*$ either as a \textit{confident} or a \textit{not-confident} prediction. Beside this first categorization, the \textit{Wiped} will also assign a classification label in the range $[1, C]$ to $x^{*}$. 

Beneath the lid, the \textit{Wiped} module is a canonical deterministic DenseNet121 model, trained on the only meaningful samples as pre-identified by the \emph{Wise}. The training procedure is the same that was described in Section~\ref{sec:uncertainty_acquisition}, with the only difference that the number of epochs is fixed and equal to 100. 

\section{Experimental Results}
\label{sec:results}
In this Section we present the experimental validation of our \textit{W2WNet}. So far, there is no agreed upon benchmark protocol to evaluate learning methods in the way they handle \emph{measurement} and \emph{labelling} noise. Therefore, we started from two well-known public datasets, the MNIST~\cite{mnist} and the CIFAR10~\cite{cifar}, and we artificially corrupted such datasets in a controlled way. By doing so, we tried to replicate different types of real-world noisy scenarios:
\begin{enumerate}[label=(\roman*)]
    \item \textit{Labelling noise (labels from a different classification task)}.
    \\In text processing, handwritten character classifications are typical mainstream tasks for CNNs. The MNIST dataset, that is made of 60000 black and white images of handwritten digits (0 to 9), was corrupted by adding a controlled percentage of alien samples randomly extracted from the EMNIST dataset~\cite{emnist}, which contains handwritten alphabetical characters. Hence, the resulting corrupted dataset, referred to as \emph{Sp-MNIST}, contains either digits (that are still the majority of the images) and letters, all with a white foreground and black background.
    By doing so, we simulate a real-world situation where a pre-processing pipeline may produce spurious samples to a downstream classifier that was specifically trained on digit classification, due to text parsing errors. This scenario is similar to any other instances of data corruption, where the spurious samples share the same characteristics of the meaningful ones in terms of color range and encoding, but belong to different classifications tasks (in this case, digits and alphabets). 
    \item \textit{Labelling and measurement noise (labels from the same classification task)}.
    \\As anticipated in Section~\ref{sec:intro}, in natural image classification, datasets may be corrupted by both labelling and measurement noise.  Mislabelling may sometimes occur due to errors during the automatic collection of a large amount of annotations from the Internet (for example, by extracting tags from the surrounding texts or keywords from search engines). On the other hand, measurement errors can always occur because of problems with acquisition and storage of the images. To simulate such scenarios, we exploited the CIFAR10 dataset, which consists of 50000 32x32 RGB images of 10 classes of natural objects. 
    As regards to labelling, the dataset was artificially corrupted by two different types of noise patterns: symmetric and pair. In the former, original labels are randomly flipped to another label. In the latter, labels are systematically flipped to the subsequent one. Both the patterns are well know in literature, as they are experienced in several image classification tasks~\cite{kohler2019uncertainty}. 
    As regards to measurement noise, we picked a random pool of images from CIFAR10 and applied three different types of transformations: (i)~blurring, via a median filter with kernel size 11; (ii)~random cropping; (iii)~random scaling. Even in this case, such image degradation is widely reported by literature, and known to be troublesome for CNN learning  in many classification tasks~\cite{dodge2016understanding}.
    As a result of our artificial corruptions, in the final dataset, referred to as \emph{Sp-CIFAR10}, a known subset of images are either given a wrong label (which, differently from the previous case, belongs to the same classification task of the original dataset), or altered in terms of image definition, scale and dynamic range. 
\end{enumerate}

To push the capabilities of our methodology to its limits, for both the above-mentioned settings, we introduced increasing amount of spurious samples (respectively, 10, 20 and 30\% of the size of the original dataset). A full characterization of the obtained validation datasets is reported in \tablename~\ref{tab:dataset}. In this table, each dataset is referred to as $[Sp]-name-[N]$, where the $Sp$ prefix indicates the presence of spurious samples, $name$ is the acronym of the original dataset and $N$ is the percentage of spurious samples with respect to the total size of the corresponding original dataset.  
\begin{table}[]
\centering
\caption{Validation benchmarks: number of images}
\label{tab:dataset}
\footnotesize
\begin{tabular}{lcccc}
\hline
\multicolumn{1}{c}{\multirow{2}{*}{\textbf{Dataset}}} & \multicolumn{2}{c}{\textbf{Train}}                                              & \multicolumn{2}{c}{\textbf{Test}}                                               \\
\multicolumn{1}{c}{}                                  & \multicolumn{1}{l}{\textbf{Meaningful}} & \multicolumn{1}{l}{\textbf{Spurious}} & \multicolumn{1}{l}{\textbf{Meaningful}} & \multicolumn{1}{l}{\textbf{Spurious}} \\ \hline
\textit{MNIST}                                          & 60000                                     & -                                      & 10000                                     & -                                      \\
\textit{$Sp-MNIST-10$}                                     & 60000                                     & 6000                                     & 10000                                     & 1000                                     \\
\textit{$Sp-MNIST-20$}                                     & 60000                                     & 12000                                    & 10000                                     & 2000                                     \\
\textit{$Sp-MNIST-30$}                                     & 60000                                     & 18000                                    & 10000                                     & 3000                                     \\
\textit{$CIFAR10$}                                        & 50000                                     & -                                      & 10000                                     & -                                      \\
\textit{$Sp-CIFAR10-10$}                                   & 50000                                     & 5000                                     & 10000                                     & 1000                                     \\
\textit{$Sp-CIFAR10-20$}                                   & 50000                                     & 10000                                    & 10000                                     & 2000                                     \\
\textit{$Sp-CIFAR10-30$}                                   & 50000                                     & 15000                                    & 10000                                     & 3000                                     \\ \hline
\end{tabular}
\end{table}
\subsection{Data cleansing capability}

As a matter of principle, our \emph{W2WNet} should follow three fundamentals: (i) if spurious samples are present, it should remove as many as possible (i.e. high sensitivity); (ii) while removing spurious samples, it should remove as little meaningful ones as possible, as they might be essential for the training of the model (i.e. high specificity); (iii) it should be able to handle datasets that do not contain any spurious samples (that is the ideal case), and leave them untouched.  

To assess all the mentioned specifications, we trained and tested our \emph{W2WNet} both on the corrupted datasets (i.e. the ones with the $Sp$ prefix in \tablename~\ref{tab:dataset}) as well as on the corresponding original ones, in the exact configuration of their reference papers.

\begin{figure}[!t]
    \centering
    \includegraphics[width=0.5\textwidth]{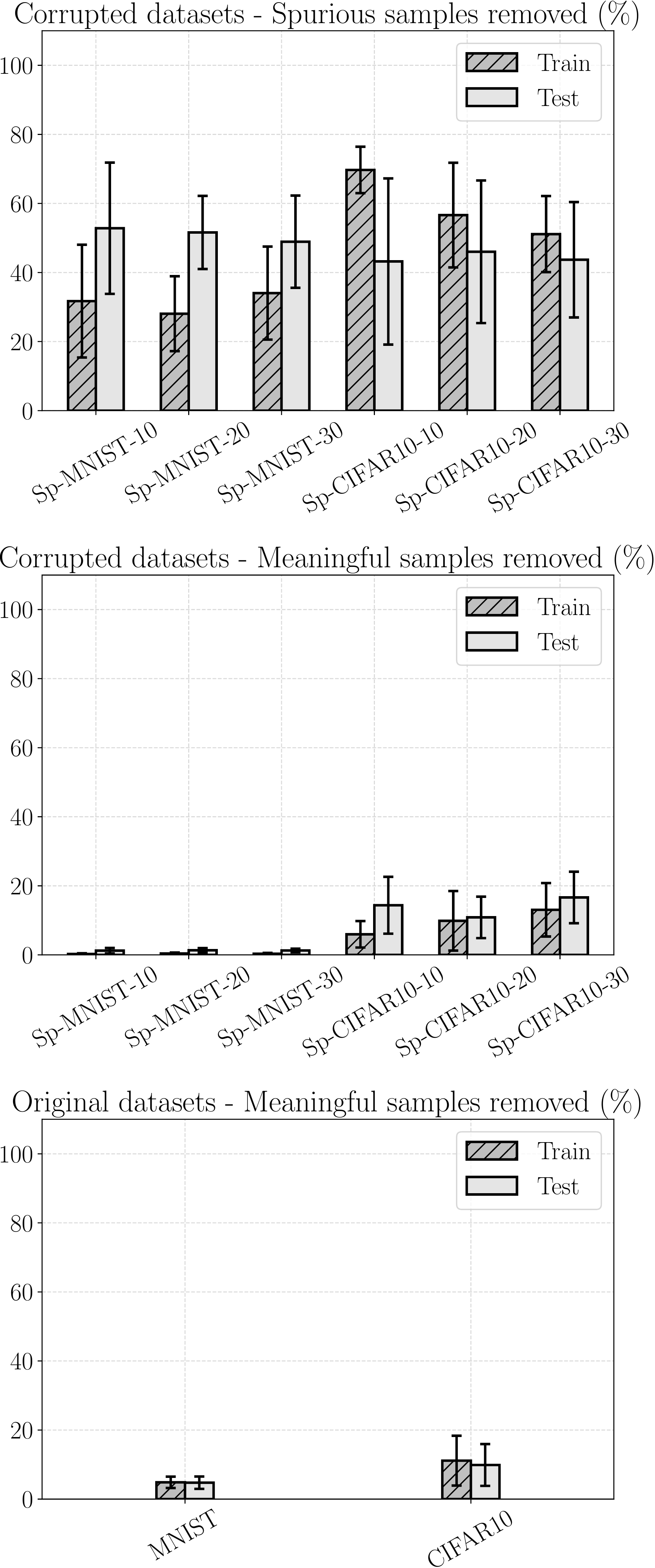}
    \caption{\emph{W2WNet} removal rates in the validation datasets. Error bars represent standard deviation of values among different classes.}
    \label{fig:removal_rates}
\end{figure}

In \figurename~\ref{fig:removal_rates} we show the results of our experiments.  Bars show the average number of images removed per dataset, separately for the training and for the test phase. In the former case, \emph{removed} images means that the model tagged them as spurious and hence removed them from the training set. In the latter case, the trained model tagged them as spurious at inference time, by providing a low-confidence prediction.

The first plot of \figurename~\ref{fig:removal_rates} reports the percentage of spurious samples which were correctly identified and removed from the corrupted datasets (i.e. the sensitivity of the model). The last two plots  report the number of meaningful samples mistakenly tagged as spurious, respectively on the corrupted datasets and on the original ones. As mentioned earlier, the lower these numbers, the higher the specificity of the model. 

As it can be gathered from the first plot, \emph{W2WNet} was able to remove at least 30\% and at best 70\% of the spurious images, when considering both the training and the test sets. Apart from the training of the $Sp-CIFAR10$, where it is possible to see a decreasing trend of the bars, the performance was quite stable at increasing number of spurious samples in the datasets. The relation between the sensitivity on the training and test sets was different for the two applications: higher on the training than on the test set for the $Sp-CIFAR10$ datasets, the opposite for the $Sp-MNIST$ ones. 

As it can be gathered from the second plot, \emph{W2WNet} proved to be reasonably specific in the corrupted datasets, removing as little as 17\% of meaningful samples in the worst case ($SP-CIFAR-30$) and almost 0\% in the best case ($SP-MNIST$). 

\begin{figure*}[!t]
    \centering
    \includegraphics[width=1\textwidth]{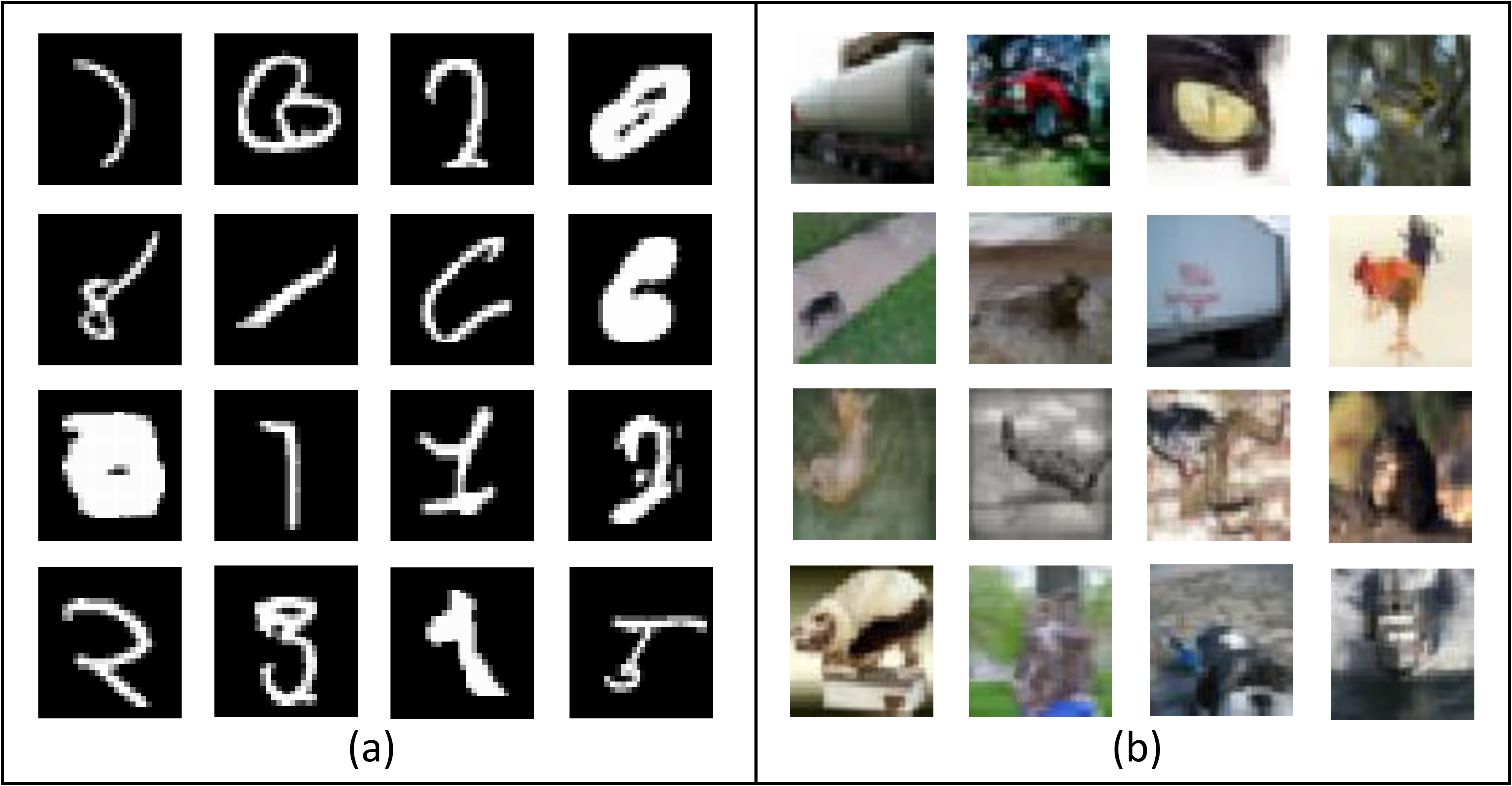}
    \caption{Examples of images tagged as \emph{spurious}, respectively from MNIST (a) and CIFAR datasets (b).}
    \label{fig:difficult_images}
\end{figure*}

Finally, by looking at the last plot, the number of meaningful images that were on average mistaken as spurious in the original datasets were 5 and 10\%, respectively in MINST and CIFAR10. A more thorough analysis revealed that in both cases these samples are very ambiguous images, that a human observer can hardly ascribe to any of the training categories (see \figurename~\ref{fig:difficult_images}). Hence, we believe that tagging such images as spurious is totally reasonable, and more importantly, it does not have a negative impact on the training, as will be showed later on.    
Overall, \emph{W2WNet} is reasonably sensitive and specific in the identification of spurious samples, and the reliability of the uncertainty measure, associated with the final prediction, is proved by our results.


\subsection{Classification performance}

At last, to assess the effectiveness of our solution in terms of positive impact on the classification performance, we compared \emph{W2WNet} against a canonical deterministic counterpart on all the datasets reported in \tablename~\ref{tab:dataset}. For this purpose we exploited a deterministic DenseNet121 model, as it is also the backbone of our \emph{W2WNet} architecture, and hence it is totally equivalent to our model in terms of depth and classification potential. For the training of the deterministic CNNs, we followed the same procedure described in Section~\ref{sec:method}, with the only difference of having set the MC dropout rate to zero. The learning rate was set to 0.1 and 0.01, respectively for the datasets derived from MNIST  and CIFAR10. 

As already anticipated in Section~\ref{sec:intro}, to the best of our knowledge, there is no published literature on deep learning methods addressing \emph{measurement} and \emph{labelling} noise coexisting together. Nonetheless, to better contextualize our validation, besides our approach and its deterministic counterpart, we also provide results obtained by representative algorithms facing either \emph{measurement} or \emph{labelling} noise. For the former category, we tested the methodology by Roy and colleagues \cite{roy2018effects}, which leverages on a not trainable low-pass filter-like CNN layer to reduce the impact of image degradation on the classification performance. For the latter, we put into effect the work by Kohler et al., in the configuration made up of a single MC droput-based classifier with 25 forward passes~\cite{kohler2019uncertainty}. For a fair comparison, both the methods were implemented using a DenseNet121 model as the backbone.

\begin{figure*}[!t]
    \centering
    \includegraphics[width=0.8\textwidth]{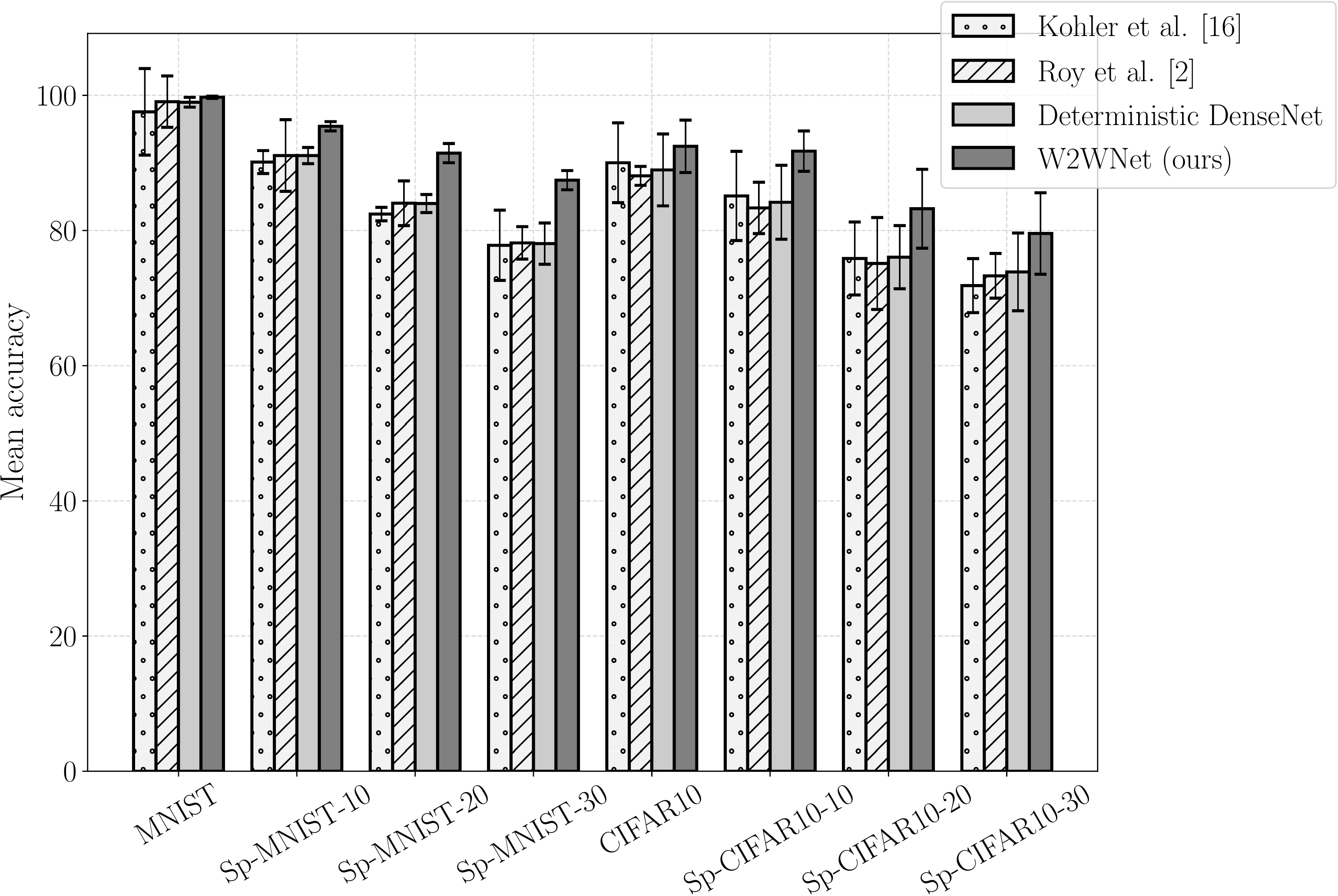}
    \caption{Mean accuracy of \emph{W2WNet} compared with representative works from literature. Error bars represent standard deviation of values among different classes.}
    \label{fig:accuracies}
\end{figure*}

The results of our experiments are reported in \figurename~\ref{fig:accuracies}, where we show the mean classification accuracy obtained by the four models (our \emph{W2WNet}, a deterministic DenseNet121, ad the two literature data cleansing approaches). As it can be observed from the plot, for all the approaches, the mean classification accuracy decreases at increasing number of spurious samples affecting the dataset (from 10 to 30 \%, see also Table \ref{tab:dataset}). This is absolutely consistent with previous literature \cite{kohler2019uncertainty}. When considering the corrupted datasets, our \emph{W2WNet} outperforms the deterministic DenseNet121 of a value between 5\% and 10\%. In addition, \emph{W2WNet} overcomes both the baseline literature solutions, which both behave similarly to DenseNet121. This is not surprising, as both methods are specifically tailored to address one type of noise solely. By a lesser margin, the accuracy of our \emph{W2WNet} was the highest even in the non-corrupted datasets.


\subsection{Real-world case study: histological images classification}
\begin{figure*}[!t]
    \centering
    \includegraphics[width=1\textwidth]{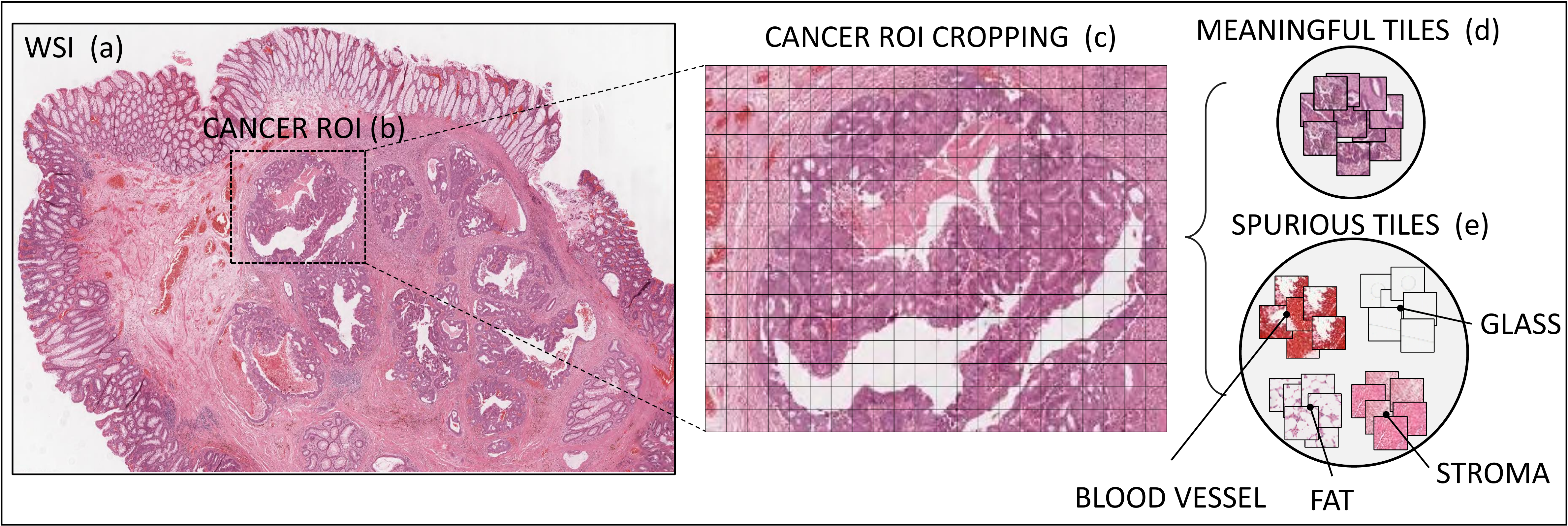}
    \caption{Generation of a digital patholohy dataset to train CNNs: typical automated procedure. (a) Whole Slide Image (WSI). (b) Identification and labelling of homogeneous Regions of Interest (ROIs). (c) Cropping ROIs into small tiles, which are all given the same label of the originating ROI. (d) Meaningful tiles (e) Spurious tiles (that is, tiles whose content is not fully representative of the given label).}
    \label{fig:roi_cropping}
\end{figure*}

Histological image analysis is the gold standard for the diagnosis and gauging of large number of cancers \cite{ponzio2019dealing}. Typically, when there is a suspicion of cancer, the patient goes through a biopsy, where a thin layer of tissue sample is resected, fixed on a slide, and stained (for example, by Hematoxylin and Eosin). Then, the pathologist analyzes the slide on the microscope looking for malignancies, which commonly cause alterations of the normal tissue architecture. The recent diffusion of digital scanners imposed the transition from standard histological slides to very large born-digital multi-resolution images called Whole-Slide Images (WSIs, see \figurename~\ref{fig:roi_cropping}(a)), whose typical size may be $100,000\times100,000$ pixels. This is rapidly changing the workflow of clinical laboratories~\cite{wsi}: the traditional visual evaluation of the samples directly under the microscope is progressively shifting to Computer-Aided Diagnosis (CAD) systems, encouraging a complete automatization of downstream image analysis. 

Recently, researchers have shown an increased interest in applying DL techniques (most often based on CNNs) to the automated assessment of the WSIs. Nonetheless, obtaining good quality training sets for the CNNs is an extremely cumbersome task, involving a number of steps: (i) manually dividing each WSI into regions of interest (ROIs), that should be homogeneous in terms of tissue architecture; (ii) manually labelling ROIs, based on the tissue category (e.g. cancer vs no-cancer, see \figurename~\ref{fig:roi_cropping}(b)); (iii) cropping ROIs into a regular grid of small tiles, that can be fed into a CNN together with their corresponding label (the same of the corresponding ROI, \figurename~\ref{fig:roi_cropping}(c)). Due to image artifacts, imprecision in the ROI delineation, or non-homogeneous content of the ROIs, the outcome of this procedure is typically a dataset which may contain a large amount of spurious tiles: that is, a significant number of tiles may have a content that is either too blurred (measurement noise) or unrelated to the label they were associated to (labelling noise), and then potentially harmful for the training of the CNN. For example, in \figurename~\ref{fig:roi_cropping}(e), a number of tiles labeled as \emph{cancer} contain a prevalence of background glass, which is obviously not meaningful to the \emph{cancer} category. This makes it a significant case-study for the exploitation of our \emph{W2WNet}.

More specifically, in our experiments we refer to the same case study described in our earlier work~\cite{ponzio}, focused on Colorectal Cancer (CRC) categorization. In this case, the classes of interest are three: (i)~Adenocarcinoma (AC), corresponding to recognizable CRC; (ii)~Tubulovillous adenoma (AD), a precursive lesion of CRC, and (iii)~Healthy tissue (H).  As detailed in \cite{ponzio}, downstream of the automated ROI cropping and labelling procedure represented in \figurename~\ref{fig:roi_cropping}, a total number of 19644 non-overlapping annotated tiles were obtained from 27 different WSIs. After ad-hoc re-examination of the tiles by a pathologist, 6144 of them were tagged as spurious, as the prevailing content of such slides (either blood vessels, adipose cells, background glass or stroma, see figure \ref{fig:roi_cropping}(e)), was not deemed meaningful to any of the three classes of interest.

For training  and  testing  purposes,  the  initial  cohort  of  27 WSIs was randomly split into two disjoint subsets (18 for training and 9 for testing), roughly balanced with respect to the classes, and then fed into our \emph{W2WNet} for data cleansing and classification. 

\begin{figure*}[!t]
    \centering
    \includegraphics[width=0.9\textwidth]{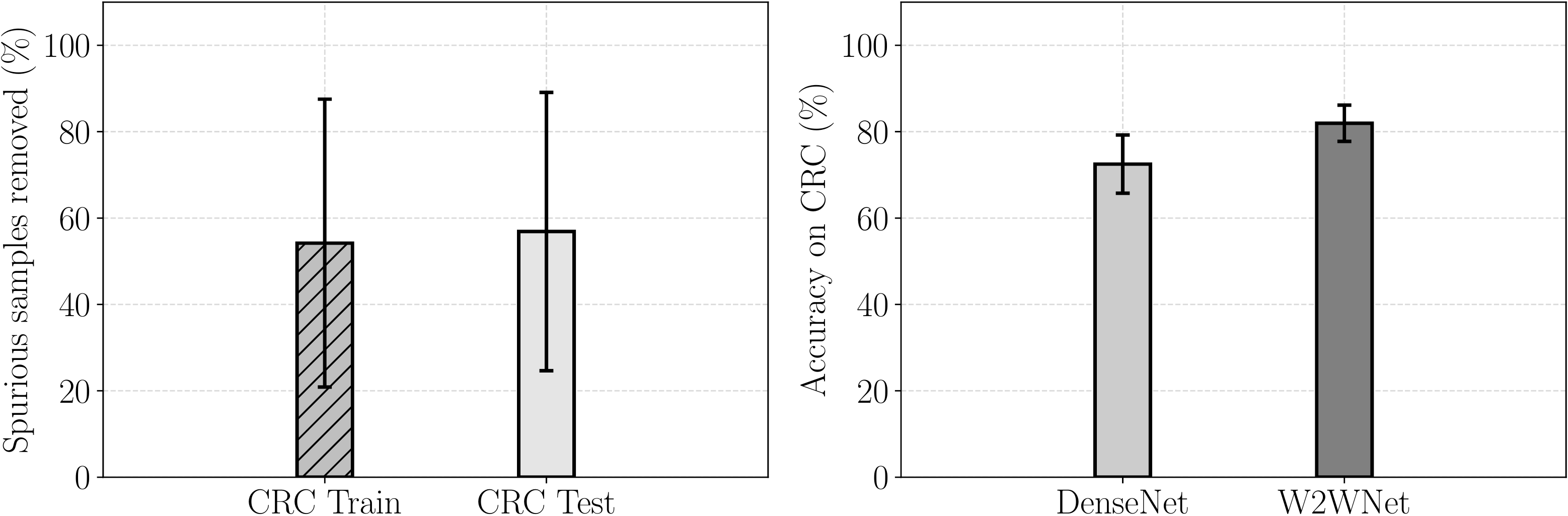}
    \caption{Removal rate of spurious samples with \emph{W2WNet} (left) and classification accuracy comparison of \emph{W2WNet} and DenseNet (right) on the CRC dataset. Error bars represent standard deviation of values among different classes.}
    \label{fig:CRC}
\end{figure*}

The results of our experiments are shown in \figurename~\ref{fig:CRC}. As it is visible from the plot on the left, our framework was able to identify 55\% and 58\% of the spurious samples from training and test set respectively. The impact on the classification is shown on the right plot, where we compare the mean classification accuracy of \emph{W2WNet} with the one obtained by the deterministic counterpart, a state-of-the-art DenseNet121 CNN, trained from scratch with learning rate set to 0.0001 and SGD optimizer. Even in this case, the accuracy of our proposed solution was higher, by about 9\% on average on the test set.


\section{Conclusions}
\label{sec:discussion}
Unfortunately, \emph{measurement} and \emph{labelling} noise are unavoidable in many real-world applications of CNNs. On one hand, the training phase of a CNN may be affected by many types of image degradation, due to problems of acquisition, encoding or storage, and mislabelling, due to faults of the manual annotation or of the automated labelling systems. On the other hand, at inference time, a CNN that was trained on a good quality dataset may be fed with low-quality images, that are completely unrelated to the ones the model was trained on. Even in such cases, a standard CNN is neither able to provide a correct prediction, nor to communicate its impossibility to provide a reliable answer. 

To address this issue, in this paper we proposed \emph{W2WNet}, a CNN architecture exploiting Bayesian probabilistic inference to i)~identify the peculiar distribution of spurious samples in a dataset, that may be affected by both measurement and labelling noise; ii)~clean the training dataset from the spurious samples and focus the learning strategy on the only meaningful ones; iii)~at inference time, provide a statistically well-founded measure of prediction confidence on the new inputs, clearly identifying the ones on which the network is too uncertain. 

Our experiments on MNIST and CIFAR10 datasets, artificially corrupted by a controlled number of spurious samples, has shown that \emph{W2WNet} can cope well with measurement and labelling noise, both in terms of sensitivity and specificity in the identification of the spurious samples. As an effect of this, \emph{W2WNet} improves on the classification accuracy of a DenseNet121 CNN, which is the deterministic counterpart of our classifier, as well as of state-of-the-art methods, which are tailored to one specific type of noise. On top of that, we found that \emph{W2WNet} outperformed the other techniques even in the classification of non-corrupted datasets (i.e. original MNIST and CIFAR10), thanks to its capability of discarding a limited number of ambiguous images from such datasets. 

Ultimately, we evaluated \emph{W2WNet} in a real-world case study from medical image analysis, that is the classification of histological samples from WSIs. Even in this case, \emph{W2WNet} was able to handle the presence of several spurious samples, that were generated by a typical dataset generation pipeline in digital pathology~\cite{ponzio}, and improve on the performance of the DenseNet121. 

In conclusion, we believe that our findings have important implications for the proficient exploitation of DL models in many real-world settings, where the presence of image quality and labelling issues typically challenge the use of classic CNN architectures, both during the training and the inference phase. 

\bibliographystyle{unsrt}
\bibliography{references.bib}  

\begin{thebibliography}{10}

\bibitem{alexnet}
Alex Krizhevsky, Ilya Sutskever, and Geoffrey~E Hinton.
\newblock Imagenet classification with deep convolutional neural networks.
\newblock In {\em Advances in neural information processing systems}, pages
  1097--1105, 2012.

\bibitem{roy2018effects}
Prasun Roy, Subhankar Ghosh, Saumik Bhattacharya, and Umapada Pal.
\newblock Effects of degradations on deep neural network architectures.
\newblock {\em arXiv preprint arXiv:1807.10108}, 2018.

\bibitem{moosavi2016deepfool}
Seyed-Mohsen Moosavi-Dezfooli, Alhussein Fawzi, and Pascal Frossard.
\newblock Deepfool: a simple and accurate method to fool deep neural networks.
\newblock In {\em Proceedings of the IEEE conference on computer vision and
  pattern recognition}, pages 2574--2582, 2016.

\bibitem{dodge2016understanding}
Samuel Dodge and Lina Karam.
\newblock Understanding how image quality affects deep neural networks.
\newblock In {\em 2016 eighth international conference on quality of multimedia
  experience (QoMEX)}, pages 1--6. IEEE, 2016.

\bibitem{dataset_labelling1}
Davood Karimi, Haoran Dou, Simon~K Warfield, and Ali Gholipour.
\newblock Deep learning with noisy labels: Exploring techniques and remedies in
  medical image analysis.
\newblock {\em Medical Image Analysis}, 65:101759, 2020.

\bibitem{dataset_labelling2}
Xin Liu, Shaoxin Li, Meina Kan, Shiguang Shan, and Xilin Chen.
\newblock Self-error-correcting convolutional neural network for learning with
  noisy labels.
\newblock In {\em 2017 12th IEEE International Conference on Automatic Face \&
  Gesture Recognition (FG 2017)}, pages 111--117. IEEE, 2017.

\bibitem{Distilling_Hilton}
Geoffrey Hinton, Oriol Vinyals, and Jeffrey Dean.
\newblock Distilling the knowledge in a neural network.
\newblock In {\em NIPS Deep Learning and Representation Learning Workshop},
  2015.

\bibitem{Chollet}
F.~{Chollet}.
\newblock Xception: Deep learning with depthwise separable convolutions.
\newblock In {\em 2017 IEEE Conference on Computer Vision and Pattern
  Recognition (CVPR)}, pages 1800--1807, 2017.

\bibitem{sun2017revisiting}
Chen Sun, Abhinav Shrivastava, Saurabh Singh, and Abhinav Gupta.
\newblock Revisiting unreasonable effectiveness of data in deep learning era.
\newblock In {\em Proceedings of the IEEE international conference on computer
  vision}, pages 843--852, 2017.

\bibitem{xiao2015learning}
Tong Xiao, Tian Xia, Yi~Yang, Chang Huang, and Xiaogang Wang.
\newblock Learning from massive noisy labeled data for image classification.
\newblock In {\em Proceedings of the IEEE conference on computer vision and
  pattern recognition}, pages 2691--2699, 2015.

\bibitem{face1}
Wilman~WW Zou and Pong~C Yuen.
\newblock Very low resolution face recognition problem.
\newblock {\em IEEE Transactions on image processing}, 21(1):327--340, 2011.

\bibitem{face2}
Chuan-Xian Ren, Dao-Qing Dai, and Hong Yan.
\newblock Coupled kernel embedding for low-resolution face image recognition.
\newblock {\em IEEE Transactions on Image Processing}, 21(8):3770--3783, 2012.

\bibitem{ullman2016atoms}
Shimon Ullman, Liav Assif, Ethan Fetaya, and Daniel Harari.
\newblock Atoms of recognition in human and computer vision.
\newblock {\em Proceedings of the National Academy of Sciences},
  113(10):2744--2749, 2016.

\bibitem{bartlett2006convexity}
Peter~L Bartlett, Michael~I Jordan, and Jon~D McAuliffe.
\newblock Convexity, classification, and risk bounds.
\newblock {\em Journal of the American Statistical Association},
  101(473):138--156, 2006.

\bibitem{Survey_label_noise}
B.~{Frenay} and M.~{Verleysen}.
\newblock Classification in the presence of label noise: A survey.
\newblock {\em IEEE Transactions on Neural Networks and Learning Systems},
  25(5):845--869, 2014.

\bibitem{kohler2019uncertainty}
Jan~M K{\"o}hler, Maximilian Autenrieth, and William~H Beluch.
\newblock Uncertainty based detection and relabeling of noisy image labels.
\newblock In {\em Proceedings of the IEEE Conference on Computer Vision and
  Pattern Recognition Workshops}, pages 33--37, 2019.

\bibitem{lakshminarayanan2017simple}
Balaji Lakshminarayanan, Alexander Pritzel, and Charles Blundell.
\newblock Simple and scalable predictive uncertainty estimation using deep
  ensembles.
\newblock In {\em Advances in neural information processing systems}, pages
  6402--6413, 2017.

\bibitem{gal2016dropout}
Yarin Gal and Zoubin Ghahramani.
\newblock Dropout as a bayesian approximation: Representing model uncertainty
  in deep learning.
\newblock In {\em international conference on machine learning}, pages
  1050--1059, 2016.

\bibitem{hendrycks2016baseline}
Dan Hendrycks and Kevin Gimpel.
\newblock A baseline for detecting misclassified and out-of-distribution
  examples in neural networks.
\newblock {\em arXiv preprint arXiv:1610.02136}, 2016.

\bibitem{kwon2020uncertainty}
Yongchan Kwon, Joong-Ho Won, Beom~Joon Kim, and Myunghee~Cho Paik.
\newblock Uncertainty quantification using bayesian neural networks in
  classification: Application to biomedical image segmentation.
\newblock {\em Computational Statistics \& Data Analysis}, 142:106816, 2020.

\bibitem{laplaceapprox}
David~JC MacKay.
\newblock {\em Bayesian methods for adaptive models}.
\newblock PhD thesis, California Institute of Technology, 1992.

\bibitem{MCMC}
Radford~M Neal.
\newblock {\em Bayesian learning for neural networks}, volume 118.
\newblock Springer Science \& Business Media, 2012.

\bibitem{variational1}
Alex Graves.
\newblock Practical variational inference for neural networks.
\newblock In {\em Advances in neural information processing systems}, pages
  2348--2356, 2011.

\bibitem{variational2}
Christos Louizos and Max Welling.
\newblock Structured and efficient variational deep learning with matrix
  gaussian posteriors.
\newblock In {\em International Conference on Machine Learning}, pages
  1708--1716, 2016.

\bibitem{srivastava2014dropout}
Nitish Srivastava, Geoffrey Hinton, Alex Krizhevsky, Ilya Sutskever, and Ruslan
  Salakhutdinov.
\newblock Dropout: a simple way to prevent neural networks from overfitting.
\newblock {\em The journal of machine learning research}, 15(1):1929--1958,
  2014.

\bibitem{rkaczkowski2019ara}
{\L}ukasz R\k{a}czkowski, Marcin Mo{\.{z}}ejko, Joanna Zambonelli, and Ewa
  Szczurek.
\newblock Ara: accurate, reliable and active histopathological image
  classification framework with bayesian deep learning.
\newblock {\em Scientific reports}, 9(1):1--12, 2019.

\bibitem{densenet}
Gao Huang, Zhuang Liu, Laurens Van Der~Maaten, and Kilian~Q Weinberger.
\newblock Densely connected convolutional networks.
\newblock In {\em Proceedings of the IEEE conference on computer vision and
  pattern recognition}, pages 4700--4708, 2017.

\bibitem{ponzio}
F.~{Ponzio}, G.~{Deodato}, E.~{Macii}, S.~{Di Cataldo}, and E.~{Ficarra}.
\newblock Exploiting “uncertain” deep networks for data cleaning in digital
  pathology.
\newblock In {\em 2020 IEEE 17th International Symposium on Biomedical Imaging
  (ISBI)}, pages 1139--1143, 2020.

\bibitem{shridhar2019comprehensive}
Kumar Shridhar, Felix Laumann, and Marcus Liwicki.
\newblock A comprehensive guide to bayesian convolutional neural network with
  variational inference.
\newblock {\em arXiv preprint arXiv:1901.02731}, 2019.

\bibitem{mnist}
Yann LeCun, L{\'e}on Bottou, Yoshua Bengio, and Patrick Haffner.
\newblock Gradient-based learning applied to document recognition.
\newblock {\em Proceedings of the IEEE}, 86(11):2278--2324, 1998.

\bibitem{cifar}
Alex Krizhevsky, Geoffrey Hinton, et~al.
\newblock Learning multiple layers of features from tiny images.
\newblock 2009.

\bibitem{emnist}
Gregory Cohen, Saeed Afshar, Jonathan Tapson, and Andre Van~Schaik.
\newblock Emnist: Extending mnist to handwritten letters.
\newblock In {\em 2017 International Joint Conference on Neural Networks
  (IJCNN)}, pages 2921--2926. IEEE, 2017.

\bibitem{ponzio2019dealing}
Francesco Ponzio, Gianvito Urgese, Elisa Ficarra, and Santa Di~Cataldo.
\newblock Dealing with lack of training data for convolutional neural networks:
  The case of digital pathology.
\newblock {\em Electronics}, 8(3):256, 2019.

\bibitem{wsi}
Navid Farahani, Anil~V Parwani, and Liron Pantanowitz.
\newblock Whole slide imaging in pathology: advantages, limitations, and
  emerging perspectives.
\newblock {\em Pathol Lab Med Int}, 7(23-33):4321, 2015.

\end{thebibliography}
\end{document}